\documentclass[10pt,twocolumn,letterpaper]{article}

\usepackage{iccv}
\usepackage{times}
\usepackage{epsfig}
\usepackage{graphicx}
\usepackage{amsmath}
\usepackage{amssymb}
\usepackage{subcaption}
\usepackage{float}
\floatstyle{ruled}
\newfloat{algorithm}{tbp}{loa}
\providecommand{\algorithmname}{Algorithm}
\floatname{algorithm}{\protect\algorithmname}

\usepackage{algorithmic}

\usepackage[pagebackref=true,breaklinks=true,letterpaper=true,colorlinks,bookmarks=false]{hyperref}

\usepackage[latin9]{inputenc}

\providecommand{\algorithmname}{Algorithm}
\floatname{algorithm}{\protect\algorithmname}

 \iccvfinalcopy 

\def\iccvPaperID{1589} 

\ificcvfinal\fi
\begin{document}

\title{The Best of Both Worlds: Combining Data-independent and Data-driven Approaches for Action Recognition  }

\author{Zhenzhong Lan, Dezhong Yao, Ming Lin, Shoou-I Yu, Alexander Hauptmann \\
{\{lanzhzh, minglin, iyu, alex@cs.cmu.edu\}, dyao@hust.edu.cn}
}

\maketitle

\begin{abstract}

Motivated by the success of data-driven convolutional neural networks (CNNs) in object recognition on static images, researchers are working hard towards developing CNN equivalents for learning video features. However,  learning video features globally has proven to be quite a challenge due to its high dimensionality, the lack of labelled data and the difficulty in processing large-scale video data. Therefore, we propose to leverage effective techniques from both data-driven and data-independent approaches to improve action recognition system. 

Our contribution is three-fold. First,  we propose a two-stream Stacked Convolutional Independent Subspace Analysis (ConvISA) architecture to show that unsupervised learning methods can significantly boost the performance of traditional local features extracted from data-independent models. Second, we demonstrate that by learning on video volumes detected by Improved Dense Trajectory (IDT), we can seamlessly combine our novel local descriptors with hand-crafted descriptors. Thus we can utilize available feature enhancing techniques  developed for hand-crafted descriptors. Finally, similar to multi-class classification framework in CNNs, we propose a training-free re-ranking technique that exploits the relationship among action classes to improve the overall performance. Our experimental results on four benchmark action recognition datasets show significantly improved performance. 
\end{abstract}

\section{Introduction}
\label{intro}

\begin{figure}
  \centering
      \includegraphics[width=0.5\textwidth]{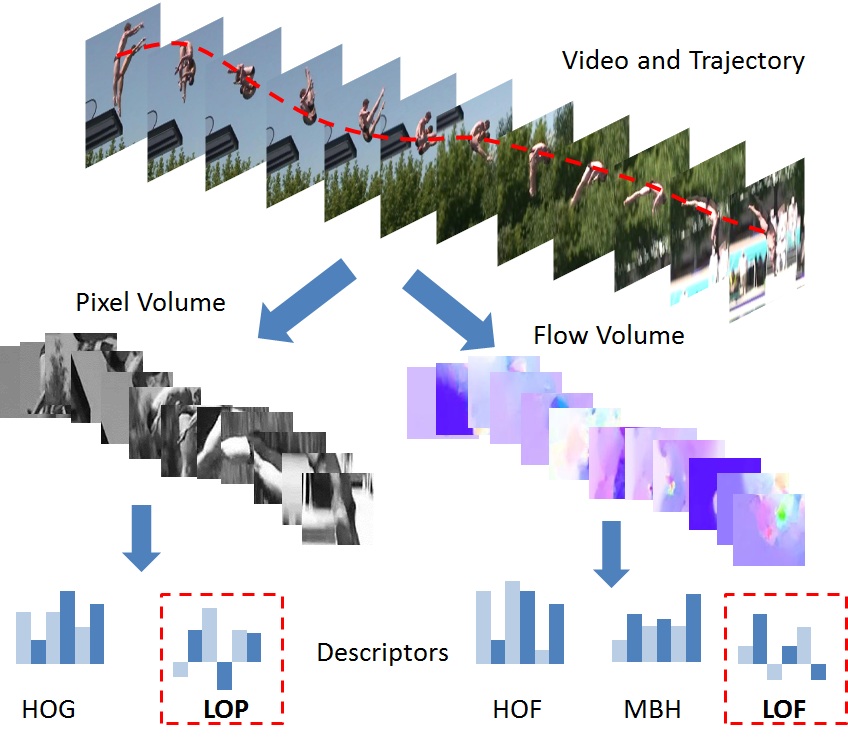}
  \caption{Illustration of our novel local video descriptors. LOP and LOF describe gray pixel and optical flow volumes, respectively. They resemble HOG/HOF/MBH in a data-driven learning framework.}
  \label{fig:illustration}
\end{figure}

Despite a long history of prior work, action recognition in videos, especially unconstrained videos that have large visual and motion variation, remains a very challenging task. Recent progress on this problem mainly relies on improvements of features, which can be categorized into two broad categories: 1) more traditional hand-crafted local features \cite{wang2009evaluation, wang2011action} and their corresponding bag-of-feature (BoF) encoding methods \cite{peng2014action}, and 2) learning based features that are mainly inspired by the success of convolutional neural networks (CNNs) for image recognition \cite{krizhevsky2012imagenet, simonyan2014two, karpathy2014large} and recurrent neural networks (RNNs) for speech recognition \cite{graves2014towards,graves2013speech, ng2015beyond}. In this paper we combine the merits of both classes of methodologies. 

Trajectory based features, especially Improved Dense Trajectories (IDT) \cite{wang2013action},  are state-of-the-art hand-crafted features that have dominated action recognition on videos in recent years. Compared with other hand-crafted motion features, IDT performs better in that it models long term motion information and has a motion boundary descriptor (MBH) that is robust to camera motion. This long-term motion information modeling, as shown in \cite{karpathy2014large, simonyan2014two}, is very hard to learn in a CNN framework. Despite its superiority, IDT still relies on simple hand-crafted local descriptors such as Histogram of Gradient (HOG) and Histogram of Optical Flow (HOF) \cite{marszalek2009actions} that took years of effort to develop.  In contrast, for image and speech recognition \cite{krizhevsky2012imagenet, ng2015beyond}, data-driven approaches have demonstrated their superiority and gradually replaced the traditional hand-crafted methods.

These revolutionary changes are largely enabled by the availability of neural networks algorithms, large scale labelled data and powerful parallel machines. Learning video features for action recognition, however, has proven to be quite a challenge due to its intrinsically high dimensionality, the lack of training data and the difficulty in processing large-scale video data \cite{karpathy2014large, simonyan2014two, ng2015beyond}. With limited training data and computational power, the learned features are generally not discriminative enough and perform worse than  IDT, especially on those datasets that have few training instances. Recent approaches \cite{simonyan2014two, ng2015beyond} circumvent these problems by learning on sampled frames or very short video clips and/or using weakly labeled data. However, video level label information can be incomplete or even missing at the frame/clip level and leads to the problem of false label assignment, which can be even worse for weakly labelled data \cite{karpathy2014large}. In other words, the imprecise frame/clip-level labels populated from video labels are too noisy for learning powerful models. With better labelled data, neural network algorithms would give superior results. Unfortunately, accurately labeling video data is very expensive. 

Though we see the value in developing fully automatically learned global video features using limited training data, we propose in this paper to revisit the traditional local feature pipeline and combine the merits of both data-independent and data-driven approaches. Inspired by the two-stream ConvNet \cite{simonyan2014two}, we introduce a two-stream Stacked Convolutional Independent Subspace Analysis (ConvISA) architecture to learn both visual appearance and motion information in an unsupervised way. As shown in Figure \ref{fig:illustration}, instead of learning on frames/short video clips, we learn on much smaller primitives--video volumes that follow the trajectories detected by IDT. The learned descriptors, called LOP (Learned descriptors of Pixel) and LOF (Learned descriptors of optical Flow), aim to improve the best performing hand-crafted descriptors within an unsupervised data-driven learning framework. The proposed architecture has several attractive properties:

\begin{itemize}
\item Compared to full video learning, small video volumes lie in a much lower dimensional space hence are computationally efficient to learn. 
\vspace{-1mm}
\item By doing unsupervised learning, we avoid the costly work of collecting labelled data and the false label assignment problem in current supervised video learning settings.  
\vspace{-1mm}
\item By learning on video volumes defined by IDT, the resulting descriptors can seamlessly combine with and boost the performance of hand-crafted descriptors. 
\vspace{-1mm}
\item By following the traditional local feature pipeline, we can easily utilize techniques developed for traditional local descriptors to improve our data-driven descriptors. 
\end{itemize}

Another merit CNN approaches have is that they naturally capture the relationships among action classes, which are hard to be captured by traditional `one-versus-all' support vector machine (SVM) framework. To address this problem, we design a Multi-class Iterative Re-ranking (MIR) method. MIR requires no training yet can significantly improve the ranking performance measured by mean average precision (MAP). To make this technique useful for improving classification accuracy, we also propose a simple ranking-score fusion technique. 

Our experimental results on four benchmark action recognition datasets show very competitive performance to the state-of-the-art results. Specifically, our methods achieve or exceed the state-of-the-art on HMDB51, Hollywood2, UCF101 and UCF50 datasets. 

In the remainder of this paper, we provide more background information about video features with an emphasis on recent attempts on learning with deep neural networks. We then describe IDT and stacked convolutional ISA in detail. After that, we evaluate our methods. Further discussions including potential improvements are given at the end.  

\section{Related Work}

In conventional video representations, features and encoding methods are the two chief reasons for considerable progress in the field. Among them, the trajectory based approaches \cite{matikainen2009trajectons,sun2009hierarchical,wang2011action,wang2013action,jiang2012trajectory}, especially the Dense Trajectory (DT) method proposed by Wang et al. \cite{wang2011action,wang2013action}, together with the Fisher Vector encoding  \cite{perronnin2010improving}  are the basis of the current state-of-the-art algorithms.  Peng et al. \cite{peng2014bag} further improved the performance of IDT by increasing the codebook sizes and fusing multiple coding methods. Sapienz \textit{et al.} \cite{sapienza2014feature} explored ways to sub-sample and generate vocabularies for Dense Trajectory features. Hoai \& Zisserman \cite{hoai2014improving} achieved the state-of-the-art on several action recognition datasets by using three useful  techniques including data augmentation, learning classifiers to model score distribution over video subsequences and learning classifiers to capture the relationship among action classes. Lan et al. \cite{lan2014beyond} proposed a multi-skip feature stacking method that combines features at multiple frame rates to achieve temporal invariance for action features. They also proposed to attach normalized 3D location information to the local descriptors and achieve good results on several action benchmark datasets. Fernando \textit{et al.} \cite{fernandomodeling} proposed to model the evolution of appearance within the video and achieved state-of-the-art results on the Hollywood2 dataset.

Independent Component Analysis \cite{hurri2003simple} (ICA) was the first approach to learn representations of videos in an unsupervised way. Le et al. \cite{le2011learning}  approached this problem using stacked ConvISA, but they only model the appearance information without using optical flow information, which has struggled to capture motion information \cite{karpathy2014large, simonyan2014two}. Generative models for understanding transformations between pairs of consecutive images are also well studied.  Recently, Ranzato et al. \cite{ranzato2014video} proposed a generative model for videos. The model uses a recurrent neural network to predict the next frame or interpolate between frames.

More recently, some success has been reported using deep CNNs for action recognition in videos. Karpathy et al. \cite{karpathy2014large} trained a deep CNNs using 1 million weakly labeled YouTube videos and reported a moderate success on using it as a feature extractor. Simonyan $\&$ Zisserman \cite{simonyan2014two} reported a result that is competitive to IDT \cite{wang2013action} by training deep CNNs using both sampled frames and stacked optical flows. Srivastava et al. \cite{srivastava2015unsupervised} tried unsupervised feature learning using long-short term memory (LSTM) but get worse results than supervised feature learning methods. 

As for modeling the relationship among classes, Bergamo \& Torresani \cite{bergamo2012meta} proposed a Meta-class method to identify related image classes based on misclassification errors on a validation set. Hou et. al. \cite{hou2014damn} proposed a method to identify similar class pairs and group them together to train `two versus rest' classifiers. By combining `two versus rest' with `one versus rest' classifiers, they observed significant improvements from baselines. Hoai \& Zisserman \cite{hoai2014improving} proposed to learn the correlation and exclusion between action classes by learning to combine the base classifier's prediction and sorted predictions from other classifiers. Unlike the aforementioned approaches that require learning and only modify the predictions once, our method is training free and iteratively updates the prediction scores given the better prediction scores in the previous iterations.

\section{Improved Dense Trajectory}

IDT improves DT feature \cite{wang2011action} by explicitly estimating camera motion and removing trajectories generated by camera motion. In this section, we first describe how trajectories are generated and then the original hand-crafted descriptors.

\subsection{Dense trajectories}
As illustrated in Figure \ref{fig:illustration}, trajectories are generated by tracking feature points across multiple frames. Gray pixel and optical flow volumes are extracted along the trajectories and represented by different kinds of descriptors. Following \cite{wang2011action}, our trajectories are extracted from multiple spatial scale with a step size of $1/\sqrt{2}$ and a scale number of 8. Feature points are densely sampled with a step of size 5 pixels and tracked in each scale separately. Specifically, each feature point
$f_t = (x_t, y_t)$ at frame t is tracked to the $t+1$ frame  by median filtering in a dense optical flow field. Mathematically, 
\begin{align}
f_{t+1} = (x_{t+1}, y_{t+1}) = (x_t, y_t) + (M * \omega)|_{(\hat{x}_t,\hat{y}_t)}
\end{align}
where M is the median filtering kernel, $(\hat{x}_t,\hat{y}_t)$ is the rounded position of $(x_t, y_t)$ and $\omega = (u_t, v_t)$ is the dense optical field calculated by the Farneback algorithm \cite{farneback2003two}. As can be seen, this tracking algorithm is computationally efficient given the optical flow field. The tracked points are put together to define a trajectory: $(f_t, f_{t+1}, f_{t+2}, . . f_{t+l-1})$, where $l$ is the length of the trajectory. Wang et al. \cite{wang2011action} experimentally show that $l=15$ gives good results across several benchmark datasets and longer trajectories may suffer from the drifting problem. Static trajectories are non-informative and hence removed.

\subsection{Hand-crafted descriptors for IDT}
There are four hand-crafted descriptors for IDT. Among them, trajectory shape, which encodes the relative locations of the trajectories, can directly get from $f_t$. The other three descriptors are Histogram of Gradient (HOG), Histogram of Optical flow (HoF) and MBH.
IDT computes HOG/HOF along the dense trajectories. For both HOG and HOF, full orientations are quantized into 8 bins with an additional zero bin for HOF. Both descriptors are normalized with their $\ell_{2}$-norm. The MBH descriptor separates the optical flow field into its x and y components. Spatial derivatives are computed for each of them and orientation information is quantized into histograms, similar to the HOG descriptor. For each component, there is a 8-bin histogram normalized by its $\ell_{2}$-norm. Since MBH represents the gradient of the optical flow, constant motion information is suppressed and only information about changes in the flow field is kept, which is a simple and effective way to remove camera motion.
These histogram-based descriptors are computed within space-time volumes aligned with a trajectory to encode the appearance and motion information. The size of the volume is $s \times s$ pixels and $l$ frames long, which corresponds to the input size of stacked ConvISA.
To embed structure information, the volume is subdivided into a spatio-temporal grid of size $s_\tau \times s_\tau \times l_\pi$.  In this paper, we fix the size of volume and grid as in IDT, i.e., $s=32$, $l=15$, $s_\tau=2$ and $l_\pi=3$ as in \cite{wang2011action}.

\section{Learning Descriptors for IDT}

Although IDT is the current state-of-the-art action recognition feature, it still relies on hand-crafted descriptors. In this section, we will describe how we use Stacked ConvISA to learn descriptors for IDT. By learning descriptors for IDT, we also aim to generalize the best performing hand-crafted features within a data-driven learning framework.

\subsection{Stacked ConvISA}
ISA is an unsupervised learning algorithm that learns features from unlabeled image patches. An ISA network \cite{hyvarinen2009natural} can be described as a two-layered network. More precisely, let matrix $W\in\mathbb{R}^{m\times n}$ and matrix $V\in\mathbb{R}^{d\times m}$ denote the parameters of the first and second layers of ISA respectively. $n$ is the dimension of the input matrix and $d$ is the dimension of outputs of ISA. $m$ is the number of latent variables between the first layer and the second layer. Typically $d\leq m\leq n$. The matrix $W$ is learned from data with orthogonal constraint $WW{}^{\top}=I$. Therefore we call $W$ the projection matrix. The matrix $V$ is given by the network structure to group the output variables of the first
layer. $V_{ij}=1$ if the $j$-th output variable of the first layer is in the $i$-th group, otherwise $V_{ij}=0$. Therefore we call $V$ the grouping matrix. Given an input pattern $X^{t}\in\mathbb{R}^{n}$, the activation of $i$-th output unit of the second layer is $p_{i}(X^{t};W,V)$
defined by 
\begin{align}
p_{i}(X^{t};W,V)\triangleq\sqrt{\sum_{k=1}^{m}V_{ik}(\sum_{j=1}^{n}W_{kj}X_{j}^{t})^{2}}\ .
\end{align}
 The ISA enforces the activation of the output unit to be sparse.
To achieve the sparse activation, it minimizes the loss function defined
on $T$ training instances:
\begin{align}
\min_{W}\quad & \sum_{t=1}^{T}\sum_{i=1}^{d}p_{i}(X^{t};W,V,)\label{eq:ISA-network-formu}\\
\mathrm{s.t.}\quad & WW{}^{\top}=I\ .\nonumber 
\end{align}
 Another way to interpret ISA is from sparse coding framework. Let
$\mathcal{G}=[\mathcal{G}_{1},\mathcal{G}_{2},\cdots,\mathcal{G}_{d}]$
denote the variable group indexes defined by $V$, that is, $j\in\mathcal{G}_{i}$
if and only if $V_{i,j}=1$. $|\mathcal{G}_i|$ defines group size, which is generally the same across groups.

As in group LASSO \cite{friedman2010note}, for
any vector $\boldsymbol{a}\in\mathbb{R}^{m}$, we defined the group
$\ell_{1}$-norm $\|\boldsymbol{a}\|_{\mathcal{G},1}$ as
\[
\|\boldsymbol{a}\|_{\mathcal{G},1}\triangleq\sum_{i=1}^{d}\sqrt{\sum_{j\in\mathcal{G}_{i}}\boldsymbol{a}_{j}^{2}}\ .
\]
 We can write $p_{i}(X^{t};W,V)$ as
\[
p_{i}(X^{t};W,V)=\|WX^{t}\|_{\mathcal{G},1}\ .
\]
 Denote $\boldsymbol{\alpha}_{t}=WX^{t}$, since $WW{}^{\top}=I$,
we have 
\[
X^{t}=W^{\dagger}\boldsymbol{\alpha}_{t}\ ,
\]
 where $W^{\dagger}$ is the Moore\textendash Penrose pseudo inverse
of $W$. Eq. (\ref{eq:ISA-network-formu}) can be re-formulated as
a sparse coding method that 
\begin{align}
\min_{W, \boldsymbol{\alpha_t}}\quad & \sum_{t=1}^{T}\|\boldsymbol{\alpha}_{t}\|_{\mathcal{G},1}\label{eq:ISA-sparse-coding-formu}\\
\mathrm{s.t.}\quad & (W^{\dagger}){}^{\top}W{}^{\dagger}=I\nonumber 
 & X^{t}=W^{\dagger}\boldsymbol{\alpha}_{t}\nonumber 
\end{align}
 Based on Eq. (\ref{eq:ISA-sparse-coding-formu}), ISA is essentially searching
a group-sparse representation $\boldsymbol{\alpha}_{t}$ of the input
signal $X_{t}$. The matrix $W^{\dagger}$ is the dictionaries of
sparse coding. The orthogonal constraint of $W^{\dagger}$ makes the
learned components maximally independent.

On high dimensional data, ISA is computationally expensive to run.  In ISA, we usually set $m=O(n)$. The optimization defined in Eq. (\ref{eq:ISA-network-formu}) will be computationally expensive for
high dimensional data since its space complexity is $O(n^{2})$ and
time complexity is $O(n^{3})$. In order to handle this problem, Le
et al. \cite{le2011learning} proposed a stacking
architecture that progressively makes use of PCA and ISA as sub-units
for unsupervised learning. The key ideas of this approach are as follows.
An ISA algorithm on small input patches was first trained. This learned
network is then used to convolve with a larger region of the input
image with a fix stride (step size). The combined responses of the
convolution step are then given as input to the next layer which is
also accomplished by another ISA algorithm with PCA as a preprocessing
step. Similar to the first layer, PCA is used to whiten the data and
reduce their dimensions such that the next layer of the ISA algorithm
only works with low dimensional inputs. The stacked model is trained
greedy layerwise in the same manner as other algorithms proposed in
the deep learning literature \cite{hinton2006fast}. Applying the
models above to the video domain is rather straightforward: the inputs
to the network are 3D video blocks instead of image patches. More
specifically, a sequence of image patches are flatten into a
vector. This vector becomes input features to the network above. The
detail algorithm of stacked ConvISA can be found in Algorithm \ref{alg:stacked-convISA}. The parameters are set as in \cite{le2011learning}. 

\begin{algorithm}
\begin{algorithmic}[1]

\STATE \textbf{Input:} Sample T trajectory volumes $X=[X^{1},X^{2},\cdots X^{t},\cdots,X^{T}]$ from videos,
where $X^{t}\in\mathbb{R}^{32\times32\times15}$.

\STATE{----- First layer network----- }

\STATE Uniformly sample T sub-volumes $\hat{X}=[\hat{X}^{1},\hat{X}^{2},\cdots,\hat{X}^{t},\cdots,\hat{X}^{T}]$
from $X$, where $\hat{X^{t}}\in\mathbb{R}^{16\times16\times5}$

\STATE Train PCA on $\hat{X}$ with PCA dimension of 300: 
\begin{align*}
\mathrm{pca\_model}_{1}= & \mathrm{PCA}_{\mathrm{train}}(\{\hat{X}^{1},\cdots,\hat{X}^{T}\},300)\\
\hat{X}_{\mathrm{pca}_{1}}= & \mathrm{PCA}_{\mathrm{apply}}(\mathrm{pca\_model}_{1},\hat{X})
\end{align*}

\STATE Train ISA on $\hat{X}_{\mathrm{pca}_{1}}$ with ISA group
size $1$: 
\begin{align*}
\mathrm{isa\_model}_{1}= & \mathrm{ISA}_{\mathrm{train}}(\hat{X}_{\mathrm{pca}_{1}},1)
\end{align*}

\STATE Convolve $X$ with first layer's model: 
\begin{align*}
X_{\mathrm{pca}_{1}}= & \mathrm{PCA}_{\mathrm{apply}}(\mathrm{pca\_model}_{1},X)\\
X_{\mathrm{ISA}_{1}}= & \mathrm{ISA}_{\mathrm{apply}}(\mathrm{isa\_model}_{1},X_{\mathrm{pca}_{1}})
\end{align*}

\STATE Using $X_{\mathrm{ISA}_{1}}$ as the output of the first layer
network : 
\[
X_{\mathrm{layer1}}=[X_{\mathrm{ISA}_{1}}]
\]

\STATE {----- Second layer network -----}

\STATE Train PCA on $X_{\mathrm{Layer1}}$ with PCA dimension of
200: 
\begin{align*}
\mathrm{pca\_model}_{2}= & \mathrm{PCA}_{\mathrm{train}}(X_{\mathrm{layer1}},200)\\
X_{\mathrm{pca}_{2}}= & \mathrm{PCA}_{\mathrm{apply}}(\mathrm{pca\_model}_{2},X_{\mathrm{layer1}})
\end{align*}

\STATE Train ISA on $X_{\mathrm{pca}_{2}}$ with group size
$2$: 
\begin{align*}
\mathrm{isa\_model}_{2}= & \mathrm{ISA}_{\mathrm{train}}(X_{\mathrm{pca}_{2}},2)\\
X_{\mathrm{ISA}_{2}}= & \mathrm{ISA}_{\mathrm{apply}}(\mathrm{isa\_model}_{2},X_{\mathrm{pca}_{2}})
\end{align*}

\STATE Stack the top 100 dimension of $X_{\mathrm{pca}_{2}}$ with
$X_{\mathrm{ISA}_{2}}$ as the output $X_{\mathrm{Layer2}}$ of the
second layer network : 
\[
X_{\mathrm{Layer2}}=[X_{\mathrm{pca}_{2}}(1:100);X_{\mathrm{ISA}_{2}}]
\]

\STATE \textbf{Output:} $X_{\mathrm{Layer2}}$

\end{algorithmic}

\protect\protect\caption{Stacked ConvISA}

\label{alg:stacked-convISA}
\end{algorithm}

\subsection{Two stream stacked ConvISA}

\begin{figure}
\centering
    \begin{subfigure}[b]{0.3\textwidth}
                \includegraphics[width=\textwidth]{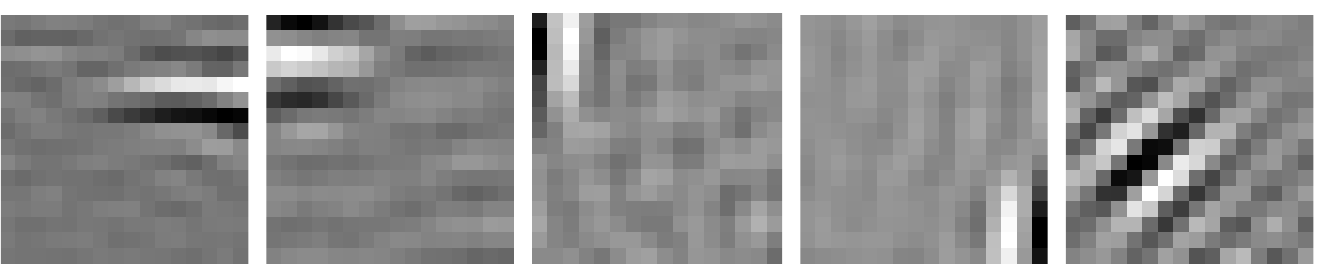}
                    \caption{Filters learned from gray pixels.}
    \end{subfigure}
    \begin{subfigure}[b]{0.3\textwidth}
                \includegraphics[width=\textwidth]{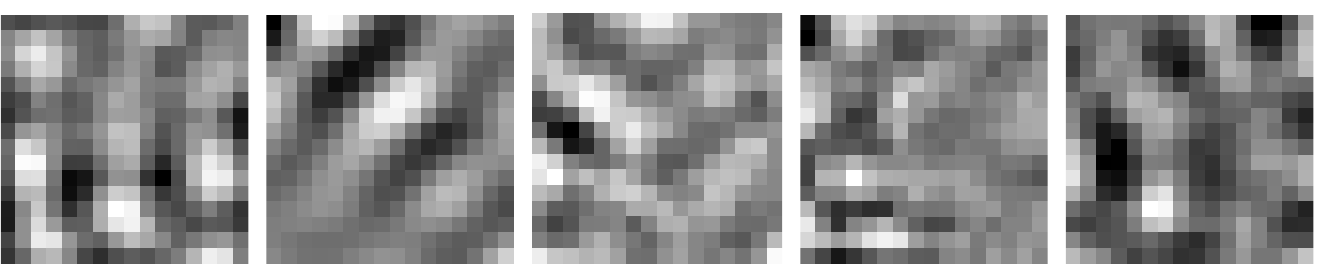}
                    \caption{Filters learned from optical flow}
    \end{subfigure}
\caption{Example filters learned using two-stream stacked ConvISA. }
\label{fig:filters}
\end{figure}

The initial stacked ConvISA was designed to learn appearance and motion together, but as shown in \cite{karpathy2014large}, spatio-temporal features learned from fixed size cuboids do not capture the motion part well \cite{simonyan2014two}. Inspired by two stream ConvNet\cite{simonyan2014two}, we propose a two-stream stacked ConvISA that learn appearance features and motion features separately. The appearance stream learns on gray scale video volumes that follow the trajectories detected by IDT and the motion stream learns to describe the corresponding optical flow volumes. In this paper, we use a two-layer stacked ConvISA as in \cite{le2011learning}. The inputs of the stacked ConvISA are  $32 \times 32 \times 15$ video volumes, within which we subsample $16 \times 16 \times 5$ video volumes to train the first layer of stacked ConvISA. Figure \ref{fig:filters} shows some randomly chosen filters learned from two stream stacked ConvISA. As can be seen, filters from optical flow have much more complex shapes than filters from gray pixels, which indicates that motion signals are more difficult to learn.

\section{Multi-class Iterative Re-ranking (MIR)}
Algorithm \ref{alg:IRR} shows our MIR algorithm. Given a score matrix $P\in\mathbb{R}^{N\times K}$ that contains K classifiers' predictions on N instances. We update each score $P_{i,j}$ iteratively by looking at other classifiers' predictions on the same instance and reducing the score using the predictions from the other classifiers. The reduction is carried out by first sorting other classifiers' predictions  $\{P_{i,1}^{(w)},P_{i,2}^{(w)},\cdots,P_{i,K}^{(w)}\}\backslash P_{i,j}^{(w)}$ in a descending order and then subtracting the weighted sum of the sorted scores from $P_{i,j}$. We use exponential decaying weight and the weighting coefficient $\alpha$ has been set to 1 throughout the paper. MIR assumes that most instances only belong to one class. We acknowledge that this assumption is quite strong in most real-world applications. However, our experiments show that even with this assumption, MIR is quite useful for improving ranking performance for real-world data where each instance can have multiple labels. We found algorithm \ref{alg:IRR} only improve the ranking performance but not classification accuracy. To make this improvement on ranking be beneficial for classification accuracy, we propose to fuse the improved ranking from MIR with the original scores. Specifically, after getting the improved ranking matrix $P_R$ from $P_{(W)}$, we normalize $P_R$ by scaling between 0 and 1 and do a z-score normalization to get $\hat{P}_R$, which becomes comparable to the original score matrix $P$. Our final prediction would be $P = \hat{P}_R + P$.

\begin{algorithm}
\begin{algorithmic}[1]

\STATE \textbf{Input:} The prediction scores of $K$ class $P\in\mathbb{R}^{N\times K}$;
Re-ranking annealing parameter $\eta$; Re-ranking weighting coefficient
$\alpha>0$; Total iteration steps $W$.

\STATE \textbf{Init:}$P^{(0)}=P$

\FOR{\inputencoding{latin1}{$w=1,2,\cdots,W-1$}\inputencoding{latin9}}

\STATE For any instance index $i\in\{1,2,\cdots,N\}$ and class index
$j\in\{1,2,\cdots,K\}$ , 
\begin{align*}
 & \Delta_{i,j}^{(w)}=\mathrm{sort}(\{P_{i,1}^{(w)},P_{i,2}^{(w)},\cdots,P_{i,K}^{(w)}\}\backslash P_{i,j}^{(w)}, \downarrow)\\
 & P_{i,j}^{(w+1)}=P_{i,j}^{(w)}- \eta^{w-1}\sum_{\substack{\substack{r=1}
\\
r\not=j
}
}^{K}\mathrm{e}^{-\alpha r}\Delta_{i,j}^{(w)} (r)
\end{align*}

\ENDFOR

\STATE\textbf{Output:} $P^{(W)}$

\end{algorithmic}

\protect\caption{Multi-class Iterative Re-ranking}
\label{alg:IRR}
\end{algorithm}

\section{Experiments}
We examine our the proposed LOP and LOF descriptors on several activity recognition datasets, predominately involving actions. The experimental results show that the new descriptors, together with several existing feature enhancing techniques and MIR, can improve IDT representations significantly. 

\subsection{Experimental setting}
As in \cite{wang2013action} and \cite{le2011learning}, IDT features are extracted using 15 frame tracking, camera motion stabilization and RootSIFT normalization and described by Trajectory, HOG, HOF, MBH, LOP and LOF descriptors. Stacked ISA models are trained on 200000 video volumes. PCA is used to reduce the dimensionality of descriptors by a factor of two. For Fisher Vector encoding,  we map the raw descriptors into a  Gaussian Mixture Model with 256 Gaussians trained from a set of randomly sampled 256000 data points. Power and $\ell_{2}$ normalization are also used before concatenating different types of descriptors into a video based representation. For classification, we use a linear SVM classifier with a fixed $C=100$ as recommended by \cite{wang2013action} and the one-versus-all approach is used for multi-class classification scenario. We use two feature enhancing techniques developed by Lan et al. \cite{lan2014beyond} to show that improvements developed for hand-crafted descriptors can also be used for our learned descriptors. These two techniques are is 3D location extension (xyt-extension) and Multi-skip Feature Stacking (MIFS). MIR is used after xyt-extension and MIFS are applied. 

\subsection{Datasets}Four representative datasets are used: The HMDB51 dataset \cite{kuehne2011hmdb} has 51 action classes and 6766 video clips extracted from digitized movies and YouTube. \cite{kuehne2011hmdb} provides both original videos and stabilized ones. We only use original videos in this paper. The Hollywood2 dataset \cite{marszalek2009actions} contains 12 action classes and 1707 video clips that are collected from 69 different Hollywood movies. We use the standard splits with training and testing videos provided by \cite{marszalek2009actions}.  The UCF101 dataset \cite{soomro2012ucf101} has 101 action classes spanning over 13320 YouTube videos clips. We use the standard splits with training and testing videos provided by \cite{soomro2012ucf101}. The UCF50 dataset \cite{reddy2013recognizing} has 50 action classes spanning over 6618 YouTube videos clips that can be split into 25 groups. The video clips in the same group are generally very similar in background. Leave-one-group-out cross-validation as recommended by \cite{reddy2013recognizing} is used. We report mean accuracy (MAcc) for HMDB51, UCF101 and UCF50 and mean average precision (MAP) for Hollyowood2 datasets as in the original papers. 

\subsection{Evaluation of our learned IDT descriptors}

\begin{table}[]
\centering
\begin{tabular}{|c|c |c | c| c | c |}
\hline
 &  HMDB51 & Hw2 & UCF101 & UCF50 \\
 \hline
 Traj& 31.9 &  42.7 & 55.2 & 69.3\\\hline
 HOG & 42.0 & 47.4& 72.4 & 77.5 \\ 
 LOP &  47.2 & 54.3 & 79.3 & 83.2\\\hline
 HOF & 49.8 &55.0 & 74.6 & 86.1 \\ 
 MBH & 52.4  & 60.8 & 81.4 & 87.0 \\ 
 LOF & 51.0 & 55.4  & 81.2 & 86.8\\\hline
 Hand crafted & 59.1 & 64.5 & 85.5 & 90.2 \\
 Traj + Learned &  56.1  & 61.1 & 85.9 & 89.3\\\hline
 Hybrid  &  \textbf{62.3} & \textbf{65.5} & \textbf{87.5} & \textbf{92.0} \\
 \hline
\end{tabular}
\caption{Comparison of our novel descriptors with hand-crafted descriptors. We report MAcc over all classes for HMDB51, UCF101 and UCF50 datasets, and MAP for Hollywood2 (Hw2). Traj means trajectory shape descriptor; Hand crafted contains four hand-crafted descriptors including Traj, HOG, HOF and MBH, Learned indicates LOP + LOF and Hybrid contains all six descriptors. }
\label{tab:descriptors}
\end{table}

In this section we compare hand-crafted descriptors with our learned descriptors as well as their combination. To compute the descriptors, we follow the parameter settings as in \cite{wang2011action} and \cite{le2011learning}. That is, $s=32, s_\tau=2, l_\pi=3$ for all descriptors. We fix the trajectory length to $l=15$. We use  $d_1 = 300$ as the dimension of first layer ISA outputs  and $d_2 = 200$ as the dimension of the second layer ISA outputs.  In section \ref{sec:parameters}, we will show that these parameters experimentally give good performance cross datasets. 

Results of the four datasets are presented in Table \ref{tab:descriptors}. Overall, our learned descriptors can boost the performance of IDT by $1\%$ to $3\%$. Comparing LOP to HOG, we can see that LOP is doing much better than HOG in capturing appearance information, resulting in $3\%$ to $8\%$ improvement. Our LOF descriptor is better than  HOF, but is still worse than MBH. These results show that even learning on video volumes defined by trajectories, motion information is hard to capture and represent. When comparing the combination of hand-crafted features and the combination of learned features, our learned features are generally worse than hand-crafted features except UCF101 datasets, but when combining all descriptors, our learned descriptors can boost the performance significantly. These improvements demonstrate that unsupervised learning methods have the potential to significantly boost the performance of hand-crafted features.

\subsection{Feature enhancing techniques}
\label{sec:feat_en}
We apply two feature enhancing techniques by Lan et al. \cite{lan2014beyond} to improve the performance of our hybrid descriptors that including hand-crafted descriptors and learned descriptors. The first technique is 3D location extension (xyt-extension), which augments the descriptors with 3D normalized location information. Another technique is Multi-skip Feature Stacking (MIFS), which stack raw features extracted from videos with different frame rates before encoding. Table \ref{tab:feauture-enhancing} summarized the results. Overall, both xyt-extension and MIFS help to improve the performance of our hybrid descriptors on all four datasets. These improvements show that by learning on local video volumes defined by IDT, we can easily utilize  available techniques developed for hand-crafted features, which would be difficult or not possible for global feature learning to do. 

\begin{table}[]
\centering
\begin{tabular}{|c|c |c | c| c |}
\hline
 &  HMDB51 & Hw2 & UCF101 & UCF50\\
 \hline
 xyt-extension& 63.9 &  67.3 & 88.2 & 94.1\\
 MIFS & 64.0 & 67.5& 88.5 & 94.5\\ 
 Combined & 66.5 & 68.8 & 89.7 & 95.1 \\ 

 \hline
\end{tabular}
\caption{ Comparison of different feature enhancing techniques for our hybrid descriptors. We report MAcc over all classes for HMDB51, UCF101 and UCF50, and MAP for Hollywood2 (Hw2).}
\label{tab:feauture-enhancing}
\end{table}

\subsection{Multi-class iterative re-ranking (MIR)}

In this section, we report results of our MIR technique on the combined results in section \ref{sec:feat_en}. Table \ref{tab:iterative} shows the number of iteration versus MAP performance on four datasets we tested. As can be seen, on all four datasets, MIR finishes within 4 iterations and converges to results that are significantly better than the baselines, which are shown in iteration 0. These improvements show that although MIR may be based on strong assumptions, it is still quite useful for real-world data. These improvements are especially seen with Hollywood2 dataset, whose source is movie data where multiple labels exist for some instances. For HMDB51, UCF101 and UCF50, after fusing the score with ranking, we get slight improvements on MAcc, from 66.5, 89.7 and 95.1 to 67.0, 90.2 and 95.4, respectively.  

\begin{table}[]
\centering
\begin{tabular}{|c|c |c | c| c |}
\hline
 Iteration&  HMDB51 & Hw2 & UCF101 & UCF50 \\
 \hline
 0& 66.8 &  68.8 & 91.2 & 96.5\\
 1 & 69.3 & 71.1& 93.4 & 97.6\\ 
 2 & 69.9 & 71.6 & 94.0 & 98.0\\ 
 3 & 70.2  & 71.9 & 94.2 & 98.1\\ 
 4 & 70.3 & 71.9 & 94.4 & 98.1\\
 5 & 70.3 & 71.9 & 94.4 & 98.1\\
 \hline
\end{tabular}
\caption{MIR iteration number versus MAP performance on four datasets we tested. Iteration 0 means the original ranking without MIR. For Hollywood2 (Hw2) and UCF50, MIR converges at 3rd iteration and for HMDB51 and UCF101, MIR converges at 4th iteration. }
\vspace{-2mm}
\label{tab:iterative}
\end{table}

\subsection{Comparison to the state-of-the-art}

\begin{table*}
\centering
\begin{tabular}{|l c |l c |l c|l c|l c|}
\hline
    \multicolumn{2}{|c|}{HMDB51 (MAcc. $\%$)} & \multicolumn{2}{|c|}{Hollywood2 (MAP $\%$)} & \multicolumn{2}{|c|}{UCF101(MAcc. $\%$)} & \multicolumn{2}{|c|}{UCF50 (MAcc. $\%$)} \\ \hline
    
Wang  \textit{et al.} \cite{wang2013action}   &57.2    &  Oneata \textit{et al.} \cite{oneata2013action} &  63.3  & Karpathy et al. \cite{karpathy2014large}  & 65.4 & Arridhana \textit{et al.} \cite{ciptadi2014movement}  & 90.0   \\
Simonyan \textit{et al.} \cite{simonyan2014two}  & 59.4 & Wang \textit{et al.} \cite{wang2013action}  &64.3    & Wang \textit{et al.} \cite{wang2013lear}  & 85.9 & Oneata \textit{et al.} \cite{oneata2013action} &90.0 \\
Peng \textit{et al.} \cite{peng2014bag}  & 61.1 & Lan \textit{et al.} \cite{lan2014beyond}  &68.0 & Peng \textit{et al.} \cite{peng2014bag}  &  87.9  & Wang \textit{et al.} \cite{wang2013action}  & 91.2  \\
Lan \textit{ et al.}  \cite{lan2014beyond} & 65.0&  Hoai \textit{et al.} \cite{hoai2014improving} & \textbf{73.6}    &Simonyan \textit{et al.}  \cite{simonyan2014two}   & 88.0    & Peng \textit{et al.} \cite{peng2014bag}  &  92.3  \\
Peng \textit{et al.} \cite{peng2014action}  & \textbf{66.8} & Fernando \textit{et al.} \cite{fernandomodeling}& \textbf{73.7} & Lan \textit{et al.} \cite{lan2014beyond} & 89.1 & Lan \textit{ et al.}  \cite{lan2014beyond} & 94.4 \\
\hline 
Ours   & \textbf{67.0} &  Ours & 71.9 & Ours & \textbf{90.2} & Ours & \textbf{ 95.4} \\
\hline
\end{tabular}
\caption{\label{tab:state-of-art}Comparison of our results to the state-of-the-art.}
\end{table*}

In Table \ref{tab:state-of-art}, we  compare our best results with the state-of-the-art approaches. From Table \ref{tab:state-of-art}, in most of the datasets, we observe improvements over the state-of-the-art. Note that although we list several most recent approaches here for comparison purposes, \textit{most of them are not directly comparable to our results due to the use of different data augmentation, features and representation methods}. For example, Hoai \& Zisserman \cite{hoai2014improving} and Fernando et al. \cite{fernandomodeling} use left-right mirrored video data augmentation to get about $2\%$ improvements on Hollywood2 dataset, which can be easily adopted by our algorithms while they can also use the MIFS and xyt extension techniques we used in this paper. The most comparable one is Lan et al.\cite{lan2014beyond}, with which we enhance our approaches. 
 Karpathy \cite{karpathy2014large} trained ConvNets on 1 million weakly labeled YouTube videos and reported  65.4\% MAcc on UCF101 datasets. Simonyan \& Zisserman \cite{simonyan2014two} reported results that are competitive to IDT by training deep convolutional neural networks using both sampled frames and optical flows and get $59.4\%$ MAcc in HMDB51 and $88.0\%$ MAcc in UCF101, which are comparable to the results of Wang \& Schmid  \cite{wang2013action}. Peng et al. \cite{peng2014action} achieves similar results to ours on the HMDB51  datasets by combining a hierarchical Fisher Vector with the original one.

\subsection{Evaluation of parameters}
\label{sec:parameters}
\begin{figure}
\centering
    \begin{subfigure}[b]{0.23\textwidth}
    \includegraphics[width=\textwidth]{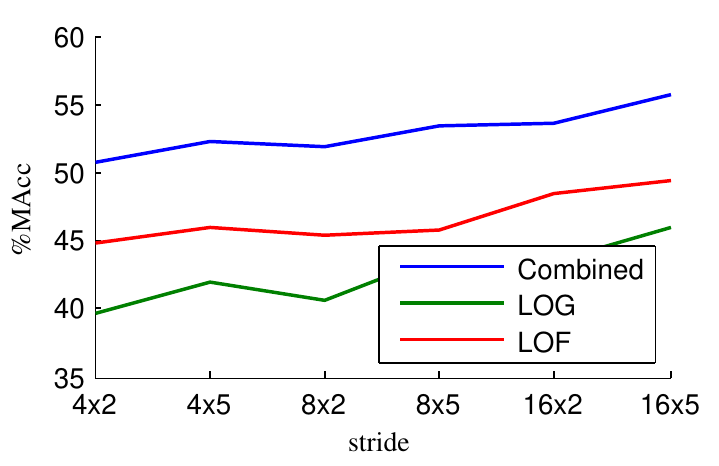}
    \caption{HMDB51}
    \end{subfigure}
    \begin{subfigure}[b]{0.23\textwidth}
     \includegraphics[width=\textwidth]{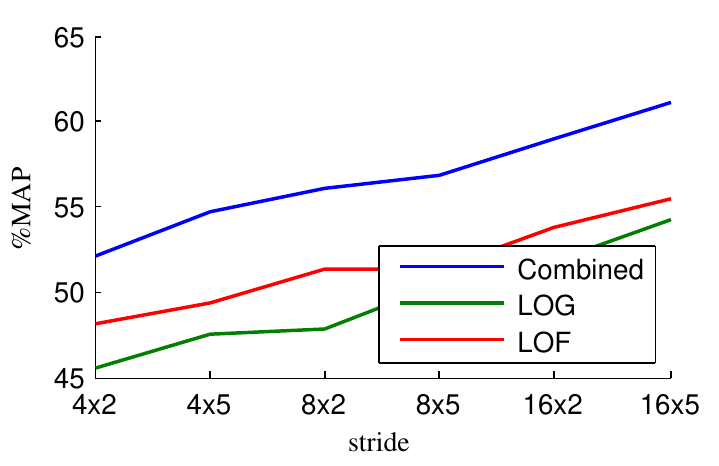}
     \caption{Hollywood2}
    \end{subfigure}
    \caption{Effect of stride on performance}
\label{fig:stride}
\vspace{-2mm}
\end{figure}

\begin{figure}
\centering
    \begin{subfigure}[b]{0.23\textwidth}
                \includegraphics[width=\textwidth]{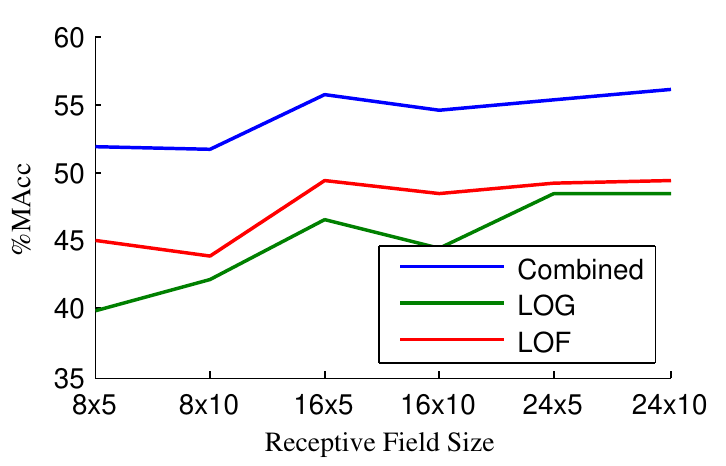}
                    \caption{HMDB51}
    \end{subfigure}
    \begin{subfigure}[b]{0.23\textwidth}
                \includegraphics[width=\textwidth]{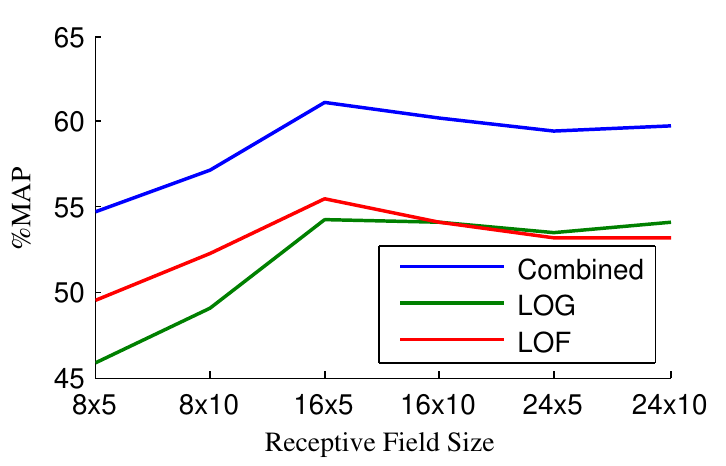}
                    \caption{Hollywood2}
    \end{subfigure}
\caption{Size of receptive field versus performance}
\label{fig:recep}
\vspace{-2mm}
\end{figure}

\begin{figure}
\centering
    \begin{subfigure}[b]{0.23\textwidth}
                \includegraphics[width=\textwidth]{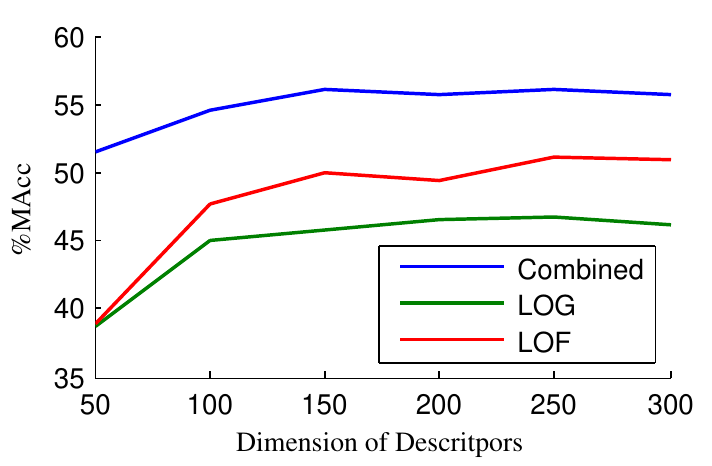}
                    \caption{HMDB51}
    \end{subfigure}
    \begin{subfigure}[b]{0.23\textwidth}
                \includegraphics[width=\textwidth]{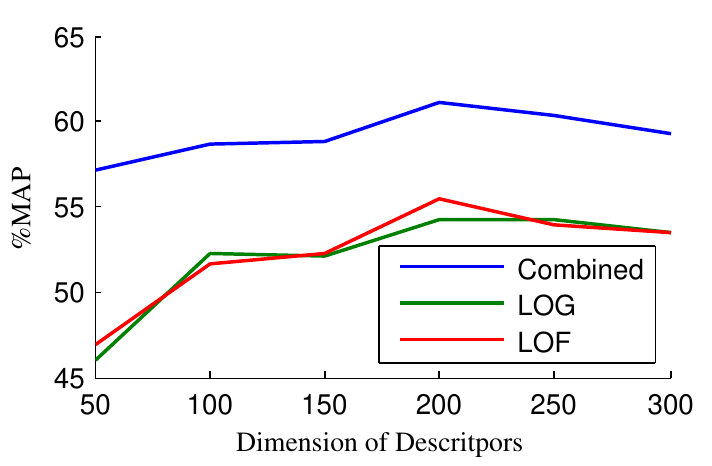}
                 \caption{Hollywood2}
    \end{subfigure}
\caption{Dimension of descriptors versus performance.}
\label{fig:dim}
\vspace{-2mm}
\end{figure}

We now move on to our characterization of performance on various axes of parameters for our new descriptors.  We report results on HMDB51 and Hollywood2, as they are the two most challenging datasets according to performance. We study the effect of stride, receptive field size (size of first layer input) and number of features (output size) as in \cite{coates2011analysis}. We evaluate the performance for a parameter at a time. The other parameters are fixed as default values.

\paragraph{Effect of stride}

First, we evaluate the effect of stride. In this experiment, we test spatial strides of 4, 8, and 16 pixels, and temporal strides of 2 and 5.  The results are summarized in Figure \ref{fig:stride}. Surprisingly, we find that, different from \cite{coates2011analysis}, larger strides consistently give better performance.  Since we observe a constant performance improvement by increasing the spatial stride from 4 to 16 and the temporal stride from 2 to 5, we attempted to increase to an even larger stride. However, the stride size is constrained by the size of the input volume. Results in Figure \ref{fig:recep} confirms that the reason for larger strides giving better results is that they cause less overlap in convolution and hence provide less redundant information to the next layer network. It is a notable fact since non-overlap scanning saves large amounts of training and prediction time. 
\vspace{-1mm}

\paragraph{Effect of receptive field}
We also study the effect of receptive field size, i.e, the input size of the first layer of the network. We test spatial sizes of 8, 16 and 24, and temporal sizes of 5 and 10 and use the maximum stride that can cover the input volumes. For example, for a respective field of $8 \times 8 \times 10 $, the stride would be $(8,  5)$. As can be seen in Figure \ref{fig:recep}, the receptive field size does not affect the performance as much as the stride. Overall, the $(16, 5)$ pair works best. Meanwhile, larger receptive fields do not give better performance. Thus, we suggest using $(16, 5)$ as it is a good balance between computational cost and performance.  
\vspace{-1mm}

\paragraph{Dimension of features}
Finally, we study the dimension of features ranging from 50 to 300 with a step size of 50. The dimension of features balances the information that is lost and the noise that has been reduced. As can be seen in Figure \ref{fig:dim}, the dimension of 200 is a good trade-off given the size of inputs we have. This result is consistent with what has been suggested by \cite{le2011learning}.

\subsection{Model generalizability}
CNN models have been shown to generalize well across different datasets \cite{razavian2014cnn}. It would also be interesting to see the generalizability of our unsupervised learned models. To show this, we apply the model learned from Hollywood2 dataset to HMDB51 dataset and vice versa. The results are shown in Table \ref{tab:gen}, from which we can see that if we train on Hollywood2 and apply it to HMDB51, there is almost no performance lost while if we train on HMDB51 and apply the model to Hollywood2, the performance drops by about 2\%. These results indicate that if we train on difficult datasets like Hollywood2, the model can generalize  well.

\begin{table}[]
\centering
\begin{tabular}{|c|c |c |}
\hline
 Train/Apply &  HMDB51 & Hollywood2 \\
 \hline
 HMDB51& 56.1 &  59.3 \\
 Hollywood2& 56.0 & 61.1\\ 
 \hline
\end{tabular}
\caption{Generalizability of Stacked ConvISA models.  }
\vspace{-2mm}
\label{tab:gen}
\end{table}

\section{Conclusions}

Different from current trend of learning video feature using deep neural networks, which is computationally intensive and label demanding, we propose in this paper to revisit the traditional local feature pipeline and combine the merits of both data-independent and data-driven approaches. As an example, we present two novel local video descriptors for IDT, a state-of-the-art local video feature. The proposed descriptors are learned using Stacked ConvISA on gray pixel and optical flow volumes that follow the trajectories detected by IDT. We also design a multi-class iterative re-ranking technique called MIR to exploit relationships among classes. Extensive experiments on four real-world datasets show that our new descriptors, when combined with hand-crafted descriptors and using MIR, achieve or exceed state-of-the-art methods. Future work would be determining the appropriate depth for Stacked ConvISA. Additionally, we would like to study other unsupervised learning methods for video feature learning.

{\small
\bibliographystyle{ieee}
\bibliography{egbib}
}

\end{document}